\documentclass[10pt,twocolumn,letterpaper]{article}

\usepackage[pagenumbers]{cvpr} % To force page numbers, e.g. for an arXiv version

\usepackage[dvipsnames]{xcolor}

\definecolor{cvprblue}{rgb}{0.21,0.49,0.74}
\usepackage[pagebackref,breaklinks,colorlinks,citecolor=cvprblue]{hyperref}
\usepackage{graphicx}
\usepackage{amsmath}
\usepackage{amssymb}
\usepackage{booktabs}
\usepackage[normalem]{ulem}
\usepackage{float}
\useunder{\uline}{\ul}{}
\def\paperID{13139} % *** Enter the Paper ID here
\def\confName{CVPR}
\def\confYear{2024}

\newcommand\blfootnote[1]{%
  \begingroup
  \renewcommand\thefootnote{}\footnote{#1}%
  \addtocounter{footnote}{-1}%
  \endgroup
}

\title{Spherical Mask: Coarse-to-Fine 3D Point Cloud Instance Segmentation with Spherical Representation}

\author{Sangyun Shin\quad Kaichen Zhou\quad Madhu Vankadari\quad Andrew Markham\quad Niki Trigoni\\Department of Computer Science \\University of Oxford\\ E-mails: \texttt{\{firstname.lastname\}@cs.ox.ac.uk}}

\begin{document}
\maketitle

\blfootnote{*Corresponding authors: Sangyun Shin, Kaichen Zhou}
\begin{abstract}
Coarse-to-fine 3D instance segmentation methods show weak performances compared to recent Grouping-based, Kernel-based and Transformer-based methods. We argue that this is due to two limitations: 1) Instance size overestimation by axis-aligned bounding box(AABB) 2) False negative error accumulation from inaccurate box to the refinement phase. In this work, we introduce \textbf{Spherical Mask}, a novel coarse-to-fine approach based on spherical representation, overcoming those two limitations with several benefits. Specifically, our coarse detection estimates each instance with a 3D polygon using a center and radial distance predictions, which avoids excessive size estimation of AABB. To cut the error propagation in the existing coarse-to-fine approaches, we virtually migrate points based on the polygon, allowing all foreground points, including false negatives, to be refined. 
During inference, the proposal and point migration modules run in parallel and are assembled to form binary masks of instances. We also introduce two margin-based losses for the point migration to enforce corrections for the false positives/negatives and cohesion of foreground points, significantly improving the performance. Experimental results from three datasets, such as ScanNetV2, S3DIS, and STPLS3D, show that our proposed method outperforms existing works, demonstrating the effectiveness of the new instance representation with spherical coordinates. The code is available at: https://github.com/yunshin/SphericalMask
\end{abstract}    
\vspace{-3mm}
\section{Introduction}
\begin{figure}[t]
  \centering
  \includegraphics[height=5.2cm,width=8.0cm]{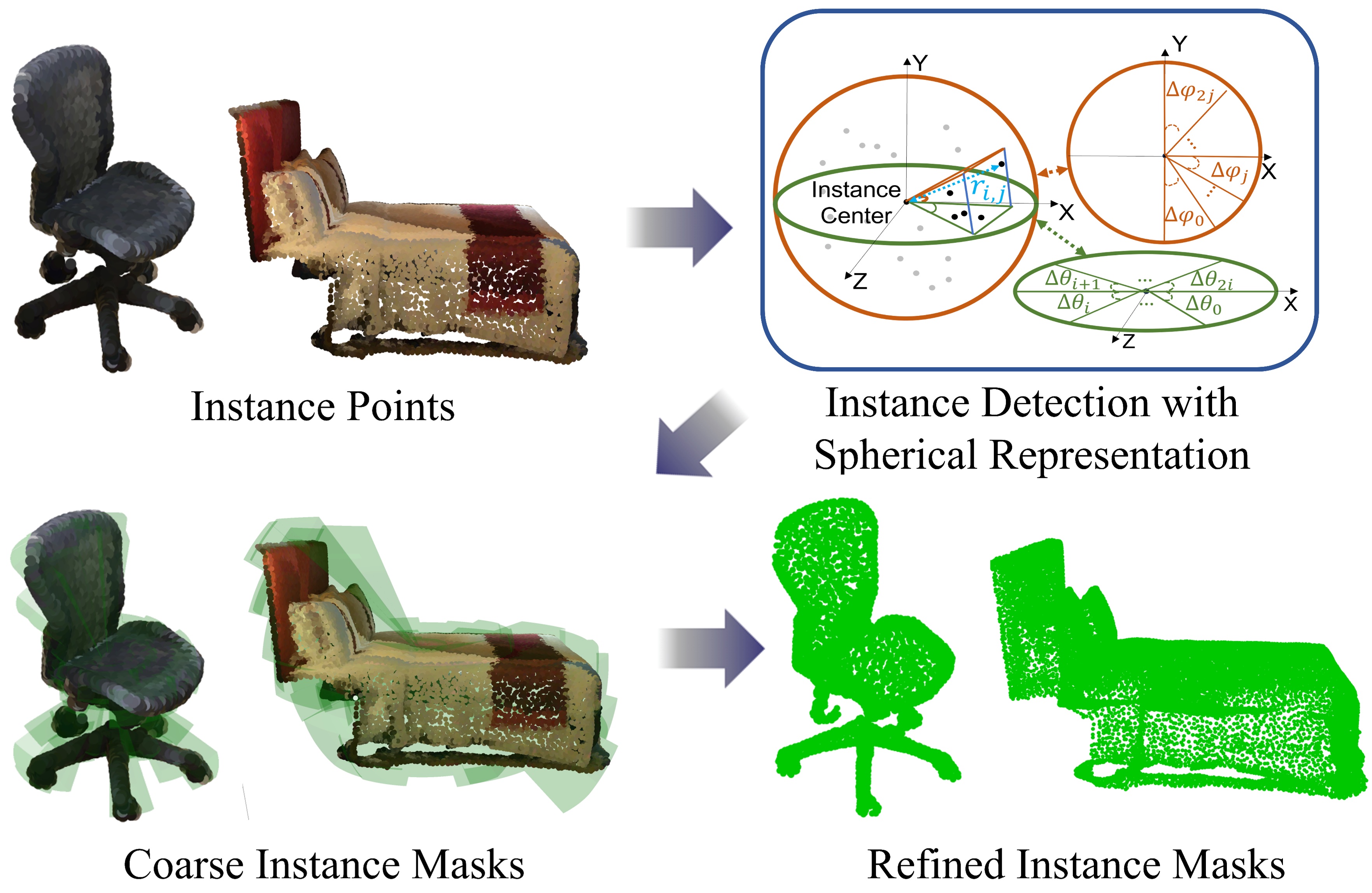}
  \caption{Pipeline of Spherical Mask with coarse-to-fine framework. Given point cloud, instances are detected with 3D polygons defined in spherical coordinates. In the refinement phase, the points virtually migrate based on the polygon to estimate fine instance masks.  }
  \vspace{-6.0mm}
  \label{fig:teaser}
\end{figure}

3D instance segmentation has gained immense attention with its wide range of applications for Indoor Scanning\cite{zhou2020joint}, Augmented Reality(AR)\cite{lehtola2017comparison}, and Autonomous Driving\cite{park2020deep}. Similar to 2D instance segmentation, the goal of the task is to identify each object along with its class label. Nevertheless, the sparse and unordered nature of point clouds has led to the development of methods different from 2D image segmentation. 

Existing approaches for 3D instance segmentation are broadly categorized into coarse-to-fine based\cite{yang2019learning,yi2019gspn,liu2020learning,hou20193d, kolodiazhnyi2023top}, grouping-based\cite{jiang2020pointgroup, chen2021hierarchical, liang2021instance, vu2022softgroup, Zhao_2023_ICCV}, kernel-based\cite{he2021dyco3d, jiang2020pointgroup, he2022pointinst3d, wu20223d, ngo2023isbnet}, and Transformer-based\cite{schult2022mask3d, sun2023superpoint, Lai_2023_ICCV, Lu_2023_ICCV, Al_Khatib_2023_ICCV} approaches. Recent progress in clustering techniques and attention mechanisms have driven the performances of the last three approaches to state-of-the-art. Compared to these three approaches, the coarse-to-fine approach has received relatively less attention due to a low accuracy, caused by two limitations: 1) False negative error propagation from the coarse detection to the refinement stage. 2) Overestimation of instance size.   

Coarse-to-fine instance segmentation first performs coarse detection, followed by refinement using the coarse detection as hard reference. The basic assumption of the approach is that the coarse detection stage always provides a neat detection for refinement. However, this assumption is often violated as the coarse detection stage cannot always produce neat outputs, which causes an issue of upper bound accuracy. For example, if the first coarse detection does not include all foreground points, the following refinement(binary classification) step has no means to include them, only possibly accumulating the error from false negative prediction. The other limitation is that the axis-aligned-bounding-box(AABB) estimation, which is typically used for the coarse detection, has been claimed to be an ill-defined problem\cite{vu2022softgroup} because commonly used box regression losses(L1, L2) result in overestimation of object sizes. For instance, the target values of AABB are minimum and maximum values in x,y,z cartesian coordinates, making the box include redundant empty space, as points only lie on the surface of the object.

In this work, we address the aforementioned two limitations of coarse-to-fine instance segmentation. Our core intuition comes from the fact that the weakness of the coarse-to-fine approach is based on the structural disentanglement of coarse detection and fine refinement phase. Instead of assuming that the coarse part should be perfect, we take a relaxation approach, which regards the coarse detection as a soft reference during the refinement phase, providing more access for refinement yet restricting access to unnecessary background points. Specifically, to improve the coarse detection part, we estimate a 3D polygon in spherical coordinates instead of AABB, alleviating the issue of excessive object size estimation. To remove the error propagation from inaccurate coarse detection to the refinement stage, we virtually migrate points based on the 3D polygon with reduced complexity in spherical coordinates. 

In summary, our method \textbf{Spherical Mask} finds each instance by estimating a 3D polygon fitting to an instance in spherical coordinates and migrating points inside or outside the polygon to produce the fine instance mask. Our contributions are:
\begin{itemize}

    \item We introduce a new alternative instance representation based on spherical coordinates, which overcomes the limitations in existing coarse-to-fine approaches.

    \item To circumvent the issue of excessive estimation of instance size, we propose \textbf{Radial Instance Detection}(\textbf{RID}) that formulates an instance into a 3D polygon as a coarse detection. 

    \item To cut the error propagation from coarse detection to the refinement phase, we introduce \textbf{Radial Point Migration(RPM)}, capable of refining both false positive and false negative points from RID.
    
    \item Extensive experiments on ScanNetV2~\cite{dai2017scannet}, S3DIS~\cite{armeni2017joint}, and STPLS3D~\cite{chen2022stpls3d} show the effectiveness of our approach, pushing the boundary of current SOTA.
    
\end{itemize}

\section{Related Work}
Existing works on point cloud instance segmentation can be categorized into proposal-based, clustering-based, kernel-based and transformer-based methods. 

\subsection{Coarse to Fine(Proposal-based) Approach}
Coarse-to-fine-based methods are built on a conceptually simple design, where they first perform coarse detection, followed by a refinement stage to acquire fine segmentation. Typically, 3D bounding box is employed for the coarse detection. 3D-SIS\cite{hou20193d} performs instance segmentation by first detecting the 3D boxes and refining points inside. 3D-BoNet\cite{yang2019learning} matches query AABBs and ground-truth instance using Hungarian algorithm for the supervision. The predicted AABBs are then concatenated with point features to produce binary masks for each instance. 
GSPN\cite{yi2019gspn} adopts set-abstraction\cite{qi2017pointnet++} to get query points and infer AABBs. The features inside the AABBs are extracted and used for per-point mask segmentation. More recently, TD3D~\cite{kolodiazhnyi2023top} proposes a fully sparse-convolutional approach for point instance segmentation. It first detects AABBs and extracts perpoint features inside the boxes to perform binary classification.
Most of the works use coarse detection(bounding box) directly as geometric features for predicting per-point binary instance masks. Thus, their accuracies greatly depend on the precision of the coarse detection, as inaccurate detection could easily lead to a large number of false negative points.

\subsection{Grouping-based Approach}
Grouping-based methods learn latent embeddings to perform per-point predictions, such as semantic categories and
clustering to acquire instances. PointGroup\cite{jiang2020pointgroup} predicts centroid offsets of each point and utilizes this shifted point cloud and original point cloud to obtain the clusters. Based on this concept, many studies improve the clustering technique with hierarchical intra-instance predictions\cite{chen2021hierarchical}, superpoint-based divisive grouping\cite{liang2021instance}, soft grouping\cite{vu2022softgroup}, and binary clustering\cite{Zhao_2023_ICCV}. The clustering-based methods have high expectations of the quality of per-point center prediction in 3D, which is challenging to generalize with various spatial extents of objects.

\subsection{Kernel-based Approach}
Kernel-based methods learn convolution kernels that aggregate point features to estimate instance masks. DyCo3D\cite{he2021dyco3d} proposes discriminative kernels by applying the clustering method from PointGroup\cite{jiang2020pointgroup}. Built on this, more recent works improved the performance by replacing the clustering part with farthest-point sampling\cite{he2022pointinst3d}, candidate localization\cite{wu20223d}, and instance-aware point sampling for the high-recall\cite{ngo2023isbnet}.

\subsection{Transformer-based Approach}
Recently, transformer-based methods have set the new SOTA. Based on mask-attention\cite{cheng2021per, cheng2022masked}, Mask3D\cite{schult2022mask3d} and SPFormer\cite{sun2023superpoint} present the pipeline that learns to output instance prediction directly from a fixed number of object queries from voxel and superpoint features, respectively. Based on these works, recent studies improve the performance with auxiliary center regression\cite{Lai_2023_ICCV}, query initialization and set grouping\cite{Lu_2023_ICCV}, spatial and semantic supervision\cite{Al_Khatib_2023_ICCV}.
Although the powerful architectural advantage has driven the performance of Transformer-based approaches, low-recall and how to distribute initial queries remain challenges.

\section{Method}

\begin{figure*}[!t]
	
	\includegraphics[height=5.2cm,width=17.7cm]{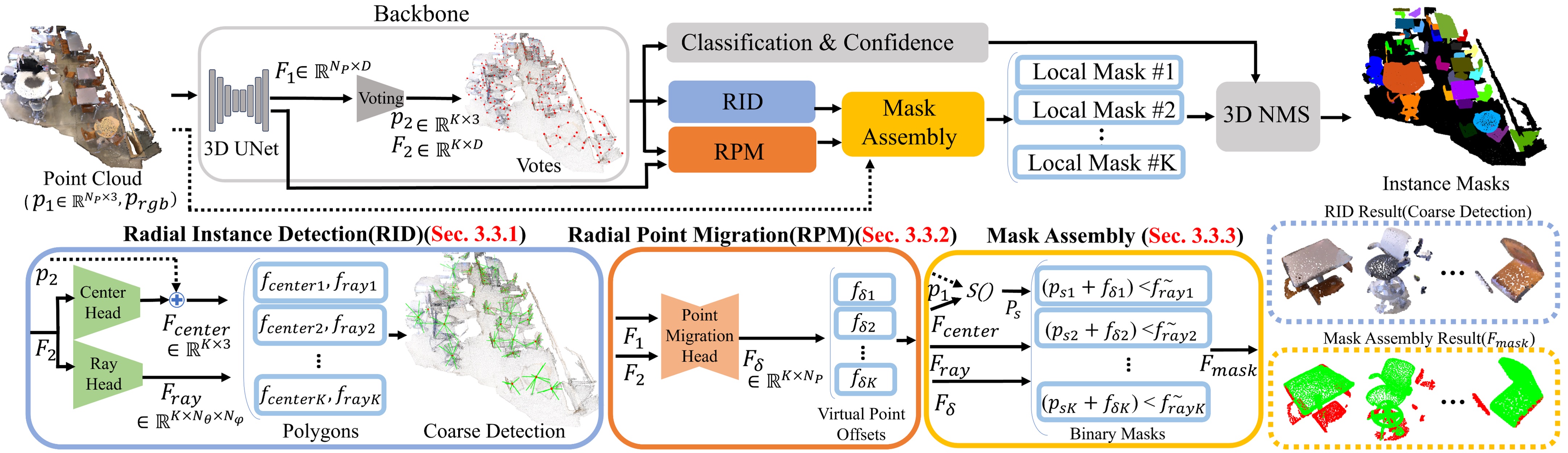}
	\caption{Overall pipeline of our proposed method based on coarse to fine approach. Given the point cloud, the backbone produces base features with 3D UNet and Voting module. Based on this, RID performs coarse detection while RPM produces the virtual point offsets to refine the coarse detection. In Mask Assembly, $K$ local binary masks are generated, where each mask is a proposal for a single instance. 3D NMS is applied to acquire the final instance masks using local binary masks, classifications, and confidence scores.}
	\label{fig:pipeline}
        \vspace{-3.0mm}
	%\vspace{-\baselineskip}
\end{figure*}
\subsection{Overview}

Given an input point cloud  $p_1 \in\mathbb{R}^{N_p \times 3}$ in 3-dimensional cartesian coordinates and the corresponding color information $p_{\text{rgb}}\in\mathbb{R}^{N_p \times 3}$, we aim to design a system that segments the point cloud into local binary masks of instances $\{o^{(i)} \in \mathbb{R}^{N_p\times 1}\}_{i=1}^{N_{o}}$ using a coarse to fine approach. Here, $N_p, N_{o}$ are the total number of points and the number of instances, respectively. This is achieved using the proposed method depicted in Figure~\ref{fig:pipeline}. Our system consists of mainly two modules: 3D backbone and proposed instance mask estimation. The details of these modules are in  Section~\ref{sec:3d_backbone} and Section~\ref{sec:spherical_masking}, respectively. %Throughout the paper, we use $p$ to refer to a point attribute and $F$ to refer to an output of a module. 

\subsection{3D backbone}
\label{sec:3d_backbone}
Our 3D backbone is similar to ~\cite{ngo2023isbnet} and is composed of two modules: a 3D encoder and a voting module. The 3D encoder
uses U-Net~\cite{ronneberger2015u} with sparse convolutions~\cite{graham20183d} to encode the given point cloud into  deep features $F_{1}\in\mathbb{R}^{N_{p}\times D}$.
%$\{f^{(j)}_1 \in \mathbb {R}^{D}\}_{j=1}^{N_p}$. 
Then, $F_1$ and the respective input points $p_1$ are fed into the voting module. The voting module performs set abstraction~\cite{qi2017pointnet++}, producing $K$ votes with query points $p_{2}\in\mathbb{R}^{K\times 3}$ and features $F_{2} \in \mathbb{R}^{K\times D}$. Please find the original paper~\cite{qi2019deep} for more details about this procedure. These votes are spread in the scene, providing features to be further processed through the proposed instance mask estimation module.
%Following~\cite{yang2019learning, ngo2023isbnet}, we adopt Hungarian algorithm to find the least-cost matching ground-truth for each vote.
The details are explained in the following sections. 

\subsection{Instance Mask Estimation}
\label{sec:spherical_masking}
In this section, we estimate the instance masks from the votes predicted in the 3D backbone section using three modules: Radial Instance Detection, Radial Point Migration, and Mask Assembly. All of the modules are explained in the following sections. For the notation simplicity, we will explain how each vote feature is processed. Therefore, we write $f_{2}$ to refer to a single vote feature of $F_{2}$ from here onwards.
\vspace{-2mm}
\subsubsection{Radial Instance Detection(Coarse Mask)} 
\begin{figure}[t]
  \centering
  \includegraphics[height=4.8cm,width=6.6cm]{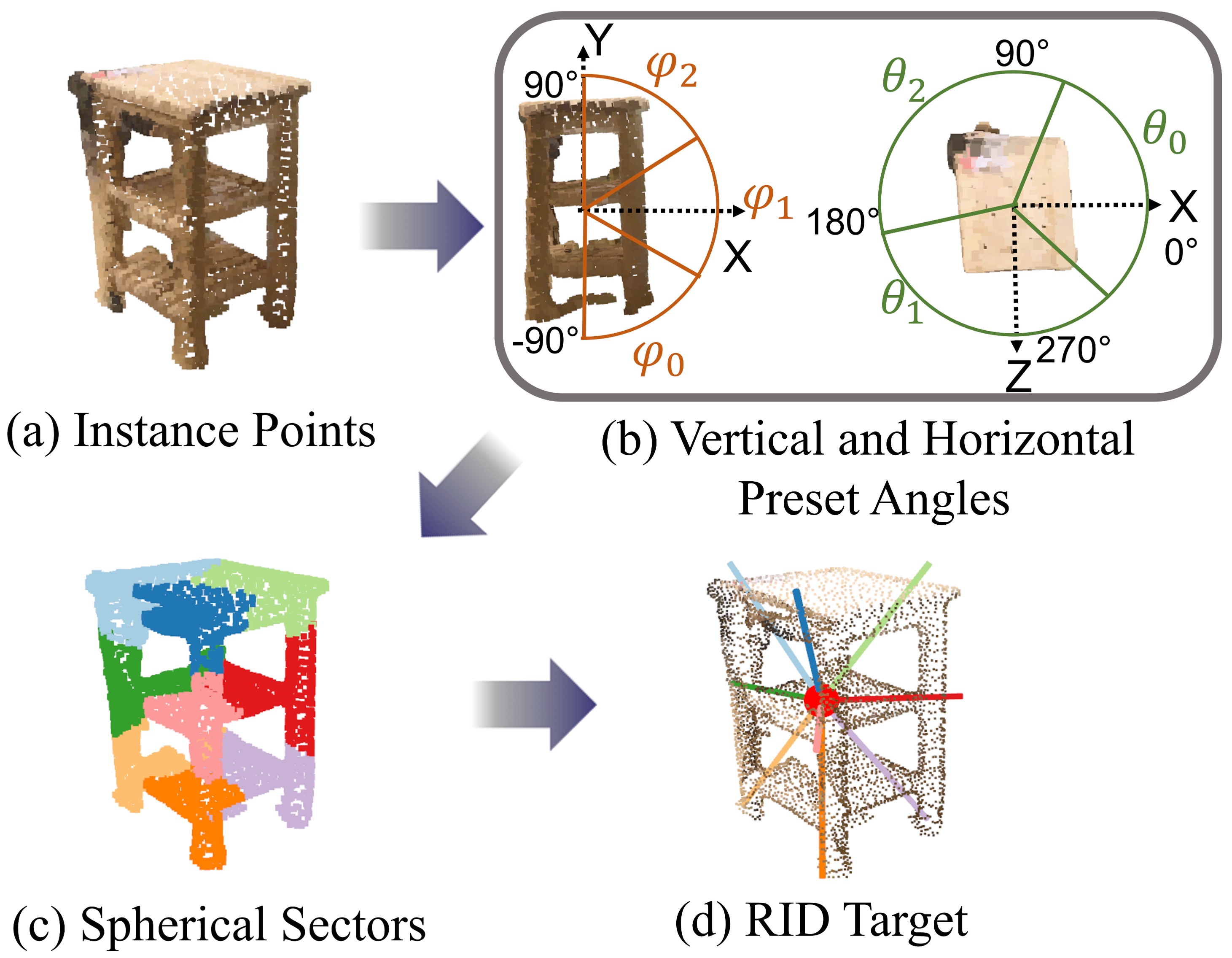}
  \vspace{-1.0mm}
  \caption{Process of RID. (a) Object points in cartesian coordinates (b) Converting points into a spherical coordinate system, using $f_{\text{center}}$, and preset angles $\theta$ and $\varphi$.  
  (c) Assigning points to each sector defined by $\theta$ and $\varphi$. The example shows 3/3 for $\theta/\varphi$. 
  (d) For each sector, the distance between the farthest point and the center becomes the target of $f_{\text{ray}}$. During inference,  
  points with smaller distance than $f_{\text{ray}}$ are considered foreground. }
  \vspace{-5.0mm}
  \label{fig:rid}
\end{figure}

Radial Instance Detection(RID) aims to detect instances for further refinement. Similar to PolarMask\cite{xie2020polarmask} for 2D segmentation, we define an instance as a 3D polygon with a center $f_{\text{center}}$ and multiple rays $f_{\text{ray}}$ emitting from the center forming each spherical sectors. Here, the sectors are defined by preset angles. Each ray then determines the distance to be considered for their corresponding sectors, as shown in Figure~\ref{fig:rid}.

We estimate the closest instances' center $f_{\text{center}}$ using offsets predicted from an MLP network, \textit{CenterHead}, which takes the respective $f_2$ as input and outputs offsets. The offsets are added to $p_{2}$ to infer $f_{center}$. After this, the input point cloud is converted into spherical coordinates using a transformation as $p_{s} = S(p_1)$ centered around $f_{\text{center}}$, where $S:(x,y,z) \rightarrow (r,\theta,\varphi)$ as follows:
\begin{equation}
    \begin{matrix}
        r = \sqrt{x^{2}+y^{2}+z^{2}},\\
        \theta = arctan \frac{x}{y},\\
        \varphi = arctan \frac{z}{\sqrt{x^{2}+y^{2}+z^{2}}}.
    \end{matrix}
\end{equation}
Here, $r$, $\theta$, $\varphi$ refer to radius, horizontal, and vertical angles in spherical coordinates, respectively. The $p_{s}$ is then divided uniformly with $N_{\theta}$ and $N_{\varphi}$ separations for horizontal and vertical direction respectively, resulting in $N_{\theta}\cdot N_{\varphi}$ sectors, as shown in Figure~\ref{fig:rid} (b) and (c). 

We consider points inside the same sector to have identical $(\theta, \varphi)$, which enables us to close the sector using a boundary estimated by $f_{\text{ray}}$. To estimate $\{f_{ray}^{(i)}\}_{i=1}^{N_{\theta}N_{\varphi}}$, another MLP named \textit{Ray Head} is employed, which takes $f_2$ as input. 

At this point, every point inside the corresponding sector's boundary from $f_{ray}$ is considered foreground. Using the boundary, RID offers tighter boundaries of instances than AABB in point cloud, as each sector is closed at the distance of the farthest foreground point in the sector, alleviating the problem of redundant space in AABB. Please refer to our supplementary material for additional visualizations.

\noindent \textbf{Coarse Instance Loss:} 
During training, the estimated $f_{\text{centre}}$ and $f_{\text{ray}}$ are compared against their respective ground truth $\textbf{g}_{\text{center}}$ and $\textbf{g}_{\text{ray}}$ to calculate $L_{\text{coarse}}$: 

\begin{equation}
    L_{\text{coarse}} = L_{\text{ray}} + L_{\text{center}},
\end{equation}
where $L_{\text{ray}}$ and $L_{\text{center}}$ are defined with L1 loss:

%\begin{equation}
%    L_{\text{ray}}=\frac{1}{N_o \times mn}\sum^{N_{o}}_{i = 1}\sum_{j = 1}^{N_{\theta} N_{\varphi}}{\left |  f_{\text{ray}}^{(\tilde{v},j)} - \textbf{g}_{\text{ray}}^{(i,j)}\right |_{1}}
%\end{equation}
%\begin{equation}
%    L_{\text{center}}=\frac{1}{N_o}\sum^{N_o}_{i}{\left |f_{\text{center}}^{\tilde{v}} - \textbf{g}_{\text{center}}^{(i)}\right |_{1}},
%\end{equation}

\begin{equation}
    L_{\text{ray}}=\frac{1}{ N_{\theta} N_{\varphi}}\sum_{i = 1}^{N_{\theta} N_{\varphi}}{\left |  f_{\text{ray}}^{(i)} - \textbf{g}_{\text{ray}}^{(i)}\right |_{1}}
\end{equation}
\begin{equation}
    L_{\text{center}}={\left |f_{\text{center}} - \textbf{g}_{\text{center}}\right |_{1}},
\end{equation}
Here, the ground-truth instance $\textbf{g}$ is matched with $f_{\text{center}}$ by an injective mapping obtained using Hungarian algorithm as ~\cite{yang2019learning, ngo2023isbnet}. For the details of the matching, please refer to Sec~\ref{sec:Training}.  
%Here, $\tilde{v} = \mathcal{H}(i)$ and $\mathcal{H}$ is an injective mapping obtained using Hungarian algorithm~\cite{yang2019learning, ngo2023isbnet} to map the $N_o$ ground truth instance to $K$ votes as $K>>N_o$. The ray target for $j_{th}$ sector 
$\textbf{g}_{\text{ray}}^{(i)}$ is set to the distance between $\textbf{g}_{\text{center}}$ and the furthest foreground point in the $i_{th}$ sector. If there are no foreground points in the sector, $\textbf{g}_{\text{ray}}^{(i)}$ is set to minimum as $1e-5$. The target center $\textbf{g}_{\text{center}}$ is calculated as mean values of foreground points of the matched groundtruth instance in cartesian coordinates.

\vspace{-2mm}
\label{Sec:RPM}
\subsubsection{Radial Point Migration(Mask Refinement)}

In this section, we introduce a refinement process to perform per-point fine-tuning. This is because the coarse detection will invariably include points belonging to the background or other instances (i.e. false positives) inside its boundary and neglect some instance points that fall outside (i.e. false negatives). We propose a conceptually simple yet effective dual that jitters individual points to belong to the correct instance. In particular, this is enabled by our innovative use of spherical coordinates - we only need to learn a single radial delta for each point to move it along the ray to the instance centroid while keeping angular quantities $\phi$ and $\theta$ constant. Note that this is a \textit{virtual} point motion - we do not alter the final point cloud. We use this as a virtual offset to obtain clean instance labels without modifying the coarse sector. 

By estimating an offset value for each point $p_{1}$, these misclassified points can be virtually migrated to being in the correct region.
Based on its good performance on per-point prediction, we adopt Dynamic Convolution in a similar manner to \cite{he2021dyco3d, ngo2023isbnet} as our \textit{Point Migration Head}, for predicting per-point offsets $F_{\delta}\in\mathbb{R}^{K\times N_{p}}$ using the vote features $F_2\in\mathbb{R}^{K\times D}$ as queries against the point features $F_1\in\mathbb{R}^{N_{p}\times D}$. For the coherence with the notations, we write $f_{\delta}\in\mathbb{R}^{N_{p}}$ to refer to an output of $F_{\delta}$, corresponding to one vote. For the learning of $f_{\delta}$, we divide points into two groups. The first is to learn the radial delta for the case of misclassification, and the second is to make instances more compact and cohesive by migrating points to the centroid of the sector. %\textcolor{blue}{It is better to say that there is only one process. As the only operation provided by the second step is to move the point based on the prediction given by the first step.}

\begin{figure}[t]
  \centering
  \includegraphics[height=2.5cm,width=8.3cm]{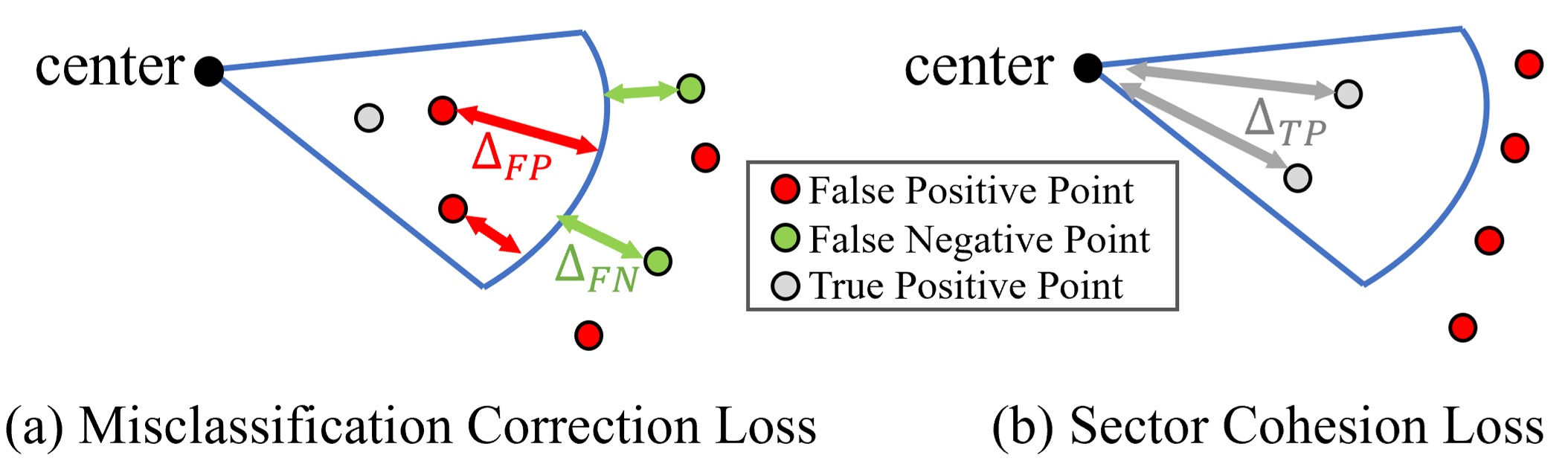}
  \vspace{-6.0mm}
  \caption{Conceptual diagram showing per-point migration following both (a)$L_{\text{mc}}$ and (b)$L_{\text{sc}}$. $\Delta_{\text{FP}}$ and $\Delta_{\text{FN}}$ are distances penalized by $L_{\text{mc}}$ with margin for misclassified points. $\Delta_{\text{TP}}$ is the distance that $L_{\text{sc}}$ penalizes to enforce the learning of general features of an instance by making each sample close to the other around the center.}
  \vspace{-4.0mm}
  \label{fig:clustreing}
\end{figure}

\vspace{5pt}
\noindent\textbf{Misclassification Correction Loss}: 
This process aims to estimate a radial delta to move the misclassified points either inside or outside the estimated coarse sector. There are two possible cases where the misclassification could occur, as shown in Figure~\ref{fig:clustreing}~(a): Instance points could lie outside the sector boundary, acting as false negatives, or background points could incorrectly lie within the sector boundary, acting as false positives. The goal is to move these points to the correct region.

Formally, given the point indices of foreground points $\{{j+}^{(i)}\}_{i=1}^{N+}$ and background points $\{{j-}^{(i)}\}_{i=1}^{N-}$ from the groundtruth, the false negative points $p_{\text{fn}}$ and the false positive points $p_{\text{fp}}$ are defined as: %\{j_{+}^{(1)}, j_{+}^{(2)}, .., j_{+}^{(N+)}\}$ 
\begin{equation}
    p_{\text{fn}} = \{p_{s}^{(j+)}: (p_{s}^{(j+)} + f_{\delta}^{(j+)}) > f_{\text{ray}}^{(\tilde{j+})}\},
\end{equation}
\begin{equation}
    p_{\text{fp}} = \{p_{s}^{(j-)}: (p_{s}^{(j-)} + f_{\delta}^{(j-)}) < f_{\text{ray}}^{(\tilde{j-})}\},
\end{equation} 

where $N+$ and $N-$ stand for the number of foreground and background points, respectively. Here, $\tilde{j}=\textit{findSector}(p_{s}^{(j)})$ where \textit{findSector(.)} is a function that takes $p_{s}^{(j)}$ as input and returns the index of the sector that $p_{s}^{(j)}$ belongs to. Please refer to the supplementary material for our implementation of \textit{findSector(.)} function.

The union of $p_{\text{fp}}$ and $p_{\text{fn}}$ forms the misclassified points $p_{\text{miss}}$ that we are interested: $p_{\text{miss}} = p_{\text{fp}}\displaystyle\cup p_{\text{fn}}$. Our aim is to push or pull them inside/outside of the ray with margins. Thus, the loss function $L_{mc}$ is formulated with soft margin loss as:
\vspace{-1mm}

%\begin{equation}
%    L_{mc} = \frac{1}{N_{\text{miss}}}\sum_{i=1}^{N_{\text{miss}}} \log (1 + \exp ( y ( p_{\text{miss}}^{(i)} + \tanh (f_{\delta}^{(\hat{i})})- f_{\text{ray}}^{(\tilde{i})} ) ))
%\end{equation}

\begin{equation}
    L_{mc} = \frac{1}{N_{\text{miss}}}\sum_{i=1}^{N_{\text{miss}}} \log (1 + \exp ( y* \tanh( p_{\text{miss}}^{(i)} + f_{\delta}^{(\hat{i})}- f_{\text{ray}}^{(\tilde{i})} ) ))
\end{equation}

%\begin{equation}
%    L^{\tilde{v}}_{mc} =  \frac{1}{\zeta N^{(\tilde{v})}_{miss}} \sum_{\eta = 1}^{\zeta} \sum_{h=1}^{N^{(\tilde{v},\eta)}_{miss}} \log\left(1 + \exp \left( y \tanh\left (\tilde{\textbf{x}}^{(\tilde{v},\eta, h)}_{miss}\right)\right)\right),
    % \sum_{j}^{|g_{\text{hard}}|} log(1+\text{exp}(y*\text{tanh}( g_{\text{hard}}^{(v,j)}- g_{\text{ray}}^{(v,\eta)})),
%\end{equation}
where 
\begin{equation}
    y=
    \begin{cases}
      \,\,\,\,1 & \text{if} \,\,\, p_{\text{miss}}^{(i)} \in p_{\text{fp}},  \\
      -1 &  \text{if} \,\,\, p_{\text{miss}}^{(i)} \in p_{\text{fn}},
    \end{cases}
\end{equation}
Here, $\tanh$ is hyperbolic-tangent function and $L_{mc}$ is calculated for all the votes that are assigned to the ground truth instance. $N_{\text{miss}}$ stands for the number of element in $p_{\text{miss}}$ and $\hat{i}$ is index of $f_{\delta}$ corresponding to $p_{\text{miss}}^{(i)}$. $f_{\text{ray}}$ is only used for reference and the gradient for learning $f_{ray}$ is not calculated.

The misclassified points around the edge,$g_{\text{ray}}$, of an instance are provided with comparably easy learning targets. In contrast, misclassified points far from the predicted rays are assigned targets with large discrepancies, encouraging larger gradients during the training.

\vspace{5pt}
\noindent\textbf{Sector Cohesion Loss}
The goal of this block is to move true-positive points to the centroid of the sector. By doing so, the sector becomes more cohesive, encouraging the learning of common and shared features of an instance as the foreground features are getting close to each other. In addition, this helps to provide a learning signal for true-positive points, as $L_{\text{mc}}$ only considers false negatives/positives. This is shown more clearly in Figure~\ref{fig:clustreing} (b).

Similar to $L_{mc}$, using point indice of foreground points $j+$, we extract true positives $p_{\text{tp}}$ as :
\begin{equation}
    p_{\text{tp}} = \{p_{s}^{(j+)}: (p_{s}^{(j+)} + f_{\delta}^{(j+)}) < f_{\text{ray}}^{(\tilde{j+})}\} 
\end{equation}
and formulate the loss with the soft margin calculation:
\vspace{-1mm}
%\begin{equation}
%    L_{sc} = \frac{1}{N_{tp}}\sum_{i=1}^{N_{tp}} \log (1 + \exp ( \tanh ( p_{tp}^{(i)} - f_{\text{center}} ) ) ),
%\end{equation}
\begin{equation}
    L_{sc} = \frac{1}{N_{\text{tp}}}\sum_{i=1}^{N_{\text{tp}}} \log (1 + \exp ( \tanh(f_{\delta}^{(\hat{i})}+ p_{\text{tp}}^{(i)} - f_{\text{center}} ))  ),
\end{equation}
where $N_{\text{tp}}$ stands for the number of element in $p_{\text{tp}}$ and $\hat{i}$ is index of $f_{\delta}$ corresponding to $p_{\text{tp}}^{(i)}$. Since $f_{\text{center}}$ is always 0 in centered spherical coordinate, the loss can be simplified as:
\vspace{-1mm}
%\begin{equation}
%    L_{sc} = \frac{1}{N_{tp}}\sum_{i=1}^{N_{tp}} \log (1 + \exp ( \tanh ( p_{tp}^{(i)}  ) ) ),
%\end{equation}
\begin{equation}
    L_{sc} = \frac{1}{N_{\text{tp}}}\sum_{i=1}^{N_{\text{tp}}} \log (1 + \exp ( \tanh ( f_{\delta}^{(i)}+p_{\text{tp}}^{(i)}   )) ),
\end{equation}
$L_{mc}$ and $L_{sc}$ together form the refinement loss $L_{\text{fine}}$ as:
\begin{equation}
    L_{\text{fine}} = L_{mc} + L_{sc}.
\end{equation} 

Our proposed virtual point migration brings three advantages for refinement over existing approaches that strictly disentangle coarse detection and refinement: 
1) Instead of only focusing on points inside the coarse detection, predicting the offsets of all points allows the refinement of even false negative points outside of $f_{\text{ray}}$, sidestepping the error accumulation from the coarse detection.
2) By considering the sector radius, $g_{\text{ray}}$, it is possible to have a soft target for each point rather than a hard target which is the center of the sector/instance. For example, false negative points outside of the sector boundary only need to be migrated a small distance to being with the sector (soft), rather than being driven towards the center (hard). This makes it easier to learn how to perform the point migration.
3) We only need to learn a one-dimensional number to migrate the point virtually along the radial line. This is far simpler than having to learn a three-dimensional offset in cartesian coordinates.

\vspace{-2mm}
\subsubsection{Mask Assembly}Our final mask is assembled by comparing virtually migrated points and $f_{ray}$. Specifically, the local binary mask $\{ f_{\text{mask}}^{(i)}\}_{i=1}^{N_{p}}$ is formed as:
\begin{equation}
    f_{\text{mask}}^{(i)} = 
    \begin{cases}
        1 & \text{if } (p_{s}^{(i)}+f_{\delta}^{(i)}) < f_{\text{ray}}^{(\tilde{i})}\\
        0 & \text{otherwise},
    \end{cases}
    \label{eq:assembly}
\end{equation}
\subsection{Training}
\label{sec:Training}

For the training, we also learn classification and confidence with respective MLPs. For classification, we apply cross-entropy loss, $L_{\text{cls}}$, for learning the classes of matched ground-truth instances. %For the confidence score of the proposal, we directly apply polar IoU loss from PolarMask\cite{xie2020polarmask}, $L_{\text{conf}}$, that measures IoU between polygons, which are defined with centers and rays emitting from the centers.
For the confidence scores of the proposals, we apply L2 loss to learn IoUs between the proposals and the groundtruth instances.
%For the confidence score of the proposal, we directly apply polar IoU loss from PolarMask\cite{xie2020polarmask}, $L_{\text{conf}}$, that measures IoU between polygons, which are defined with centers and rays emitting from the centers.
Similar to~\cite{ngo2023isbnet}, we also duplicate the number of grountruth for 4 times and create a cost matrix $C$:
\begin{equation}
    C(k,i) = L_{\text{coarse}}(k,i) + L_{\text{fine}}(k,i) + L_{\text{cls}}(k,i),
\end{equation}
where $L(,)$ refers to the loss value calculated using $k_{th}$ vote and $i_{th}$ ground-truth instance. Referring $C$, we apply Hungarian algorithm to find the least-cost injective mapping from each ground-truth instance to the votes. The final loss using the acquired ground-truths is:
\begin{equation}
    \vspace{-1mm}
    L =  \lambda_{\text{1}}L_{\text{cls}} + \lambda_{\text{2}}L_{\text{conf}} + \lambda_{\text{3}}L_{\text{coarse}} + \lambda_{\text{4}}L_{\text{fine}} 
    \vspace{-1mm}
\end{equation}

\section{Experiment}
\begin{table*}[h]
\scriptsize
\centering
\begin{tabular}{p{1.9cm}|p{0.3cm}p{0.6cm}|p{0.3cm}p{0.3cm}p{0.3cm}p{0.3cm}p{0.3cm}p{0.3cm}p{0.3cm}p{0.3cm}p{0.3cm}p{0.3cm}p{0.3cm}p{0.3cm}p{0.3cm}p{0.3cm}p{0.3cm}p{0.3cm}p{0.3cm}p{0.3cm}}

\hline
Method      & $\text{mAP}$           & $\text{mAP}_{50}$         &\rotatebox[origin=c]{90}{bath} & \rotatebox[origin=c]{90}{bed}  & \rotatebox[origin=c]{90}{bk.shf} & \rotatebox[origin=c]{90}{cabinet} & \rotatebox[origin=c]{90}{chair} & \rotatebox[origin=c]{90}{counter} & \rotatebox[origin=c]{90}{curtain} & \rotatebox[origin=c]{90}{desk} & \rotatebox[origin=c]{90}{door} & \rotatebox[origin=c]{90}{other} & \rotatebox[origin=c]{90}{picture} & \rotatebox[origin=c]{90}{fridge} & \rotatebox[origin=c]{90}{s. cur.} & \rotatebox[origin=c]{90}{sink} & \rotatebox[origin=c]{90}{sofa} & \rotatebox[origin=c]{90}{table} & \rotatebox[origin=c]{90}{toilet} & \rotatebox[origin=c]{90}{wind.} \\ \hline
3D-BoNet(C)\cite{yang2019learning}    & 25.3          & 48.8          & 51.9 & 32.4 & 25.1   & 13.7    & 34.5  & 3.1     & 41.9   & 6.9  & 16.2 & 13.1  & 5.2     & 20.2   & 33.8    & 14.7 & 30.1 & 30.3  & 65.1   & 17.8  \\
TD3D(C)\cite{kolodiazhnyi2023top}        & 48.9          & 75.1          & 85.2 & 51.1 & 43.4   & 32.2    & 73.5  & 10.1    & 51.2   & 35.5 & 34.9 & 46.8  & 28.3    & 51.4   & 67.6    & 26.8 & 67.1 & 51.0  & 90.8   & 32.9  \\ 
SoftGroup(G)\cite{vu2022softgroup}   & 50.4          & 76.1          & 66.7 & 57.9 & 37.2   & 38.1    & 69.4  & 7.2     & 67.7   & 30.3 & 38.7 & 53.1  & 31.9    & 58.2   & 75.4    & 31.8 & 64.3 & 49.2  & 90.7   & 38.8  \\
PBNet(G)\cite{Zhao_2023_ICCV}       & 57.3          & 74.7          & 92.6 & 57.5 & 61.9   & 47.2    & 73.6  & 23.9    & 48.7   & 38.3 & 45.9 & 50.6  & 53.3    & 58.5   & 76.7    & 40.4 & 71.7 & 55.9  & 96.9   & 38.1  \\ 
DKNet(K)\cite{wu20223d}       & 53.2          & 71.8          & 81.5 & 62.4 & 51.7   & 37.7    & 74.9  & 10.7    & 50.9   & 30.4 & 43.7 & 47.5  & 58.1    & 53.9   & 77.5    & 33.9 & 64.0 & 50.6  & 90.1   & 38.5  \\
ISBNet(K)\cite{ngo2023isbnet}      & 55.9          & 75.7          & 93.9 & 65.5 & 38.3   & 42.6    & 76.3  & 18.0    & 53.4   & 38.6 & 49.9 & 50.9  & 62.1    & 42.7   & 70.4    & 46.7 & 64.9 & 57.1  & 94.8   & 40.1  \\ 
MAFT(T)\cite{Lai_2023_ICCV}        & 57.8          & 78.6          & 88.9 & 72.1 & 44.8   & 46.0    & 76.8  & 25.1    & 55.8   & 40.8 & 50.4 & 53.9  & 61.6    & 61.8   & 85.8    & 48.2 & 68.4 & 55.1  & 93.1   & 45.0  \\
QueryFormer(T)\cite{Lu_2023_ICCV} & {\ul 58.3}          & {\ul 78.7}          & 92.6 & 70.2 & 39.3   & 50.4    & 73.3  & 27.6    & 52.7   & 37.3 & 47.9 & 53.4  & 53.3    & 69.7   & 72.0    & 43.6 & 74.5 & 59.2  & 95.8   & 36.3  \\ \hline
Ours(C)           & \textbf{61.6} & \textbf{81.2} & 94.6 & 65.4 & 55.5   & 43.4    & 76.9  & 27.1    & 60.4   & 44.7 & 50.5 & 54.9  & 69.8    & 71.6   & 77.5    & 48.0 & 74.7 & 57.5  & 92.5   & 43.6  \\ \hline
\end{tabular}
\vspace{-3mm}
\caption{Quantitative comparison of top-performing methods for each approach on ScanNetV2 \textbf{hidden} test set. (C),(G),(K), and (T) next to the names of the methods refer to coarse-to-fine, grouping, kernel, and Transformer based methods, respectively. All the methods take the same input, such as point cloud and corresponding color information. The best results are in \text{bold}, and the second best ones are in {\ul underlined}.}
\label{tab:scannet_test}

\end{table*}

\subsection{Dataset}
%\noindent \textbf{Dataset}
We evaluate our method on three datasets: ScanNetV2~\cite{dai2017scannet}, S3DIS~\cite{armeni2017joint}, STPLS3D~\cite{chen2022stpls3d}. Following are descriptions of each dataset.

\begin{figure*}[h!]

	\includegraphics[height=6.5cm,width=17.3cm]{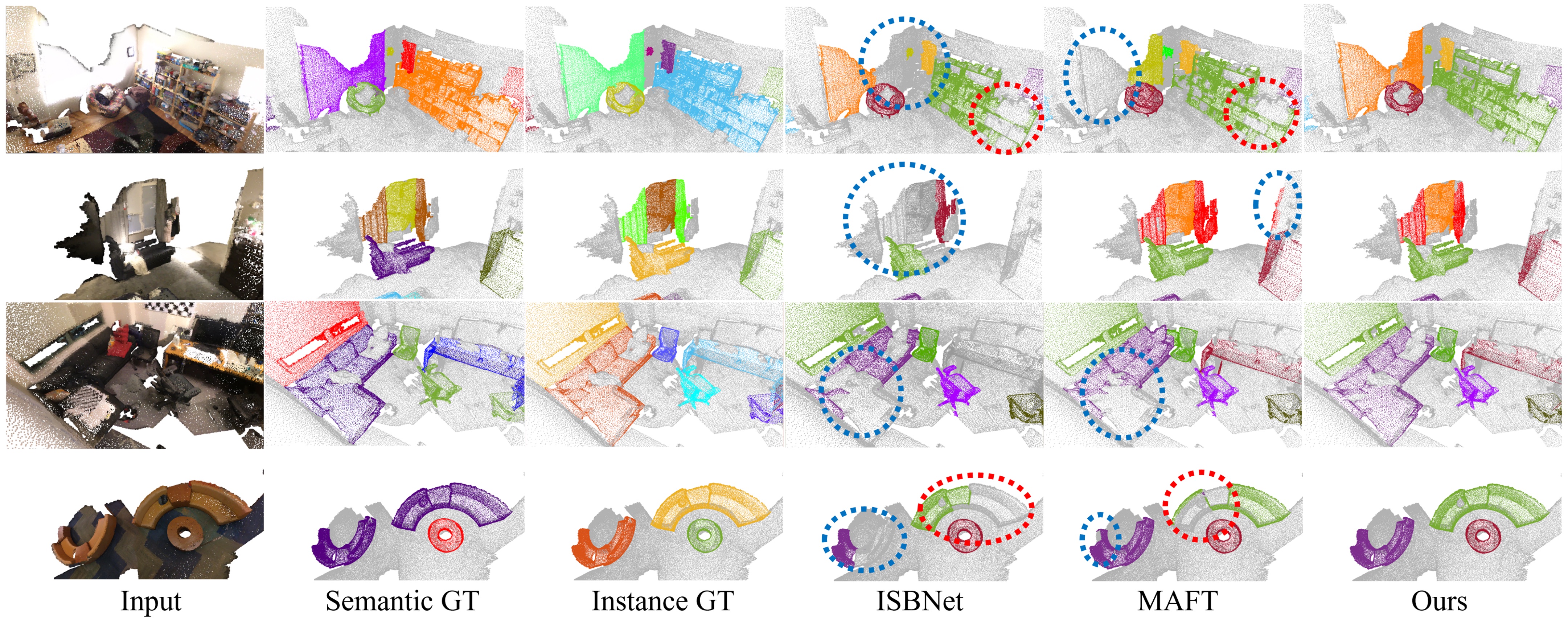}
        \vspace{-3mm}
	\caption{Qualitative comparison of ISBNet\cite{ngo2023isbnet}, MAFT\cite{Lai_2023_ICCV}, and ours on ScanNetV2 validation set.}
	\label{fig:qualitative}
	%\vspace{-\baselineskip}
\end{figure*}

\begin{table}[]
\centering
\small
\begin{tabular}{c|cccc}
\hline
Method         & Venue  & mAP           & AP50          & AP25          \\ \hline
3D-SIS(C)\cite{hou20193d}      & CVPR19 & -             & 18.7          & 35.7          \\
GSPN(C)\cite{yi2019gspn}        & CVPR19 & -             & 37.8          & 53.4          \\
TD3D(C)\cite{kolodiazhnyi2023top}        & WACV23 & 47.3          & 71.2          & 81.9          \\ \hline
PointGroup(G)\cite{jiang2020pointgroup}  & CVPR20 & 34.8          & 51.7          & 71.3          \\
SSTNet(G)\cite{liang2021instance}      & ICCV21 & 49.4          & 64.3          & 74.0          \\
MaskGroup(G)\cite{zhong2022maskgroup}   & ICME22 & 27.4          & 42.0          & 63.3          \\
SoftGroup(G)\cite{vu2022softgroup}   & CVPR22 & 46.0          & 67.6          & 78.9          \\
RPGN(G)\cite{dong2022learning}        & ECCV22 & -             & 64.2          & -             \\
PBNet(G)\cite{Zhao_2023_ICCV}       & ICCV23 & 54.3          & 70.5          & 78.9          \\ \hline
PointInst3D(K)\cite{he2022pointinst3d} & EECV22 & 45.6          & 63.7          &               \\
DKNet(K)\cite{wu20223d}       & ECCV22 & 50.8          & 66.9          & 76.9          \\
ISBNet(K)\cite{ngo2023isbnet}      & CVPR23 & 54.5          & 73.1          & 82.5          \\ \hline
Mask3D(T)\cite{schult2022mask3d}      & ICRA23 & 55.2          & 73.7          & 82.9          \\
3IS-ESSS(T)\cite{Al_Khatib_2023_ICCV}    & ICCV23 & 56.1          & 75.0          & 83.7          \\
QueryFormer(T)\cite{Lu_2023_ICCV} & ICCV23 & 56.5          & 74.2          & 83.3          \\
MAFT(T)\cite{Lai_2023_ICCV}        & ICCV23 & {\ul 58.4}          & {\ul 75.6}          & {\ul 84.5}          \\ \hline
Ours(C)        & -      & \textbf{62.3} & \textbf{79.9} & \textbf{88.2} \\ \hline
\end{tabular}
\vspace{-3mm}
\caption{Quantitative 3D instance segmentation results on ScanNetV2 validation set. (C),(G),(K), and (T) next to the names of the methods refer to coarse-to-fine, grouping, kernel, and Transformer based methods, respectively. The best results are in \text{bold}, and the second best ones are in {\ul underlined}.}
\label{tab:scannetv2_val}
\vspace{-2mm}
\end{table}

\noindent \textbf{ScanNetV2} ScanNetV2 dataset consists of 1201, 312, and 100 scans with 18 object
classes for training, validation, and testing, respectively. We
report the evaluation results on the validation and hidden test sets as in the existing works. 

\noindent \textbf{S3DIS}
S3DIS dataset contains 271 scenes from 6 areas with 13 categories. We report evaluations for both Area 5. Additional evaluation results with 6-fold cross-validation can be found in the supplementary material.

\noindent \textbf{STPLS3D}
The STPLS3D dataset is an aerial photogrammetry point cloud dataset from real-world and synthetic environments. It
includes 25 urban scenes of 6$km^{2}$ and 14 instance
categories. Following ~\cite{chen2021hierarchical, vu2022softgroup, ngo2023isbnet}, we use scenes 5, 10, 15, 20, and 25 for validation and the rest for training.

\begin{table}[]
\small

\centering
\begin{tabular}{ccccc}
\hline
Method                           & mAP  & $\text{AP}{\text{50}}$ & $\text{mPrec}_{\text{50}}$ & $\text{mRec}_{\text{50}}$ \\ \hline
\multicolumn{1}{c|}{GSPN~\cite{yi2019gspn} }       & -    & -     & 36.0          & 28.7         \\
\multicolumn{1}{c|}{PointGroup~\cite{jiang2020pointgroup}}  & -    & 5.78  & 61.9          & 62.1         \\
\multicolumn{1}{c|}{HAIS~\cite{chen2021hierarchical}}        & -    & -     & 71.1          & 65.0         \\
\multicolumn{1}{c|}{SSTNet~\cite{liang2021instance}}      & 42.7 & 59.3  & 65.6          & 64.2         \\
\multicolumn{1}{c|}{SoftGroup~\cite{vu2022softgroup}}   & 51.6 & 66.1  & 73.6          & 66.6         \\
\multicolumn{1}{c|}{Mask3D~\cite{schult2022mask3d}}      & 56.5 & 69.3  & 68.7          & 70.7         \\
\multicolumn{1}{c|}{RPGN~\cite{dong2022learning}}        & -    & -     & 64.0          & 63.0         \\
\multicolumn{1}{c|}{PointInst3D~\cite{he2022pointinst3d}} & -    & -     & {\ul 73.1}          & 65.2         \\
\multicolumn{1}{c|}{DKNet~\cite{wu20223d}}       & -    & -     & 70.8          & 65.3         \\
\multicolumn{1}{c|}{ISBNet~\cite{ngo2023isbnet}}      & 56.3 & 67.5  & 70.5          & 72.0         \\
\multicolumn{1}{c|}{PBNet~\cite{Zhao_2023_ICCV}}       & 53.5 & 66.4  & \textbf{74.9}          & 65.4         \\
\multicolumn{1}{c|}{QueryFormer~\cite{Lu_2023_ICCV}} & {\ul 57.7} & {\ul 69.9}  & 70.5          & {\ul 72.2}         \\
\multicolumn{1}{c|}{MAFT~\cite{Lai_2023_ICCV}}  & -    & 69.1  & -             & -            \\ \hline
\multicolumn{1}{c|}{Ours}        & \textbf{60.5} & \textbf{72.3}  & 71.3          & \textbf{76.3} \\ \hline      

\end{tabular}
\vspace{-3mm}
\caption{Quantitative 3D instance segmentation results on S3DIS Area 5. The best results are in \text{bold}, and the second best ones are in {\ul underlined}.}
\label{tab:s3dis}
\vspace{-2mm}
\end{table}

\subsection{Evaluation Metric}
We adopt average precision as our primary evaluation metric. Average precision is extensively used in vision tasks such as
object detection and instance segmentation tasks. The metric calculates precisions by varying the IoU threshold. Following the existing works, we evaluate our model with three IoU thresholds: $AP$, $AP_{50}$, $AP_{25}$. $AP_{50}$ and $AP_{25}$ stand for average precisions with IoU threshold as $25\%$ and $50\%$, respectively. $AP$ is an averaged score by varying IoU thresholds from $50\%$ to $95\%$ by increasing the threshold with step size $5\%$.
For S3DIS, we also evaluate our model with mean precision (mPrec50), and mean
recall with IoU threshold as $50\%$(mRec50).

\subsection{Implementation Detail}
We build our model on PyTorch framework~\cite{paszke2019pytorch} and train it for 300
epochs with AdamW optimizer with a single NVIDIA A10 GPU. The batch size is set to 10. The learning rate and weight decay are initialized to 0.001 and 0.0001.
Cosine annealing~\cite{zhang2021point} is used for scheduling the learning rate. Following~\cite{vu2022softgroup,ngo2023isbnet},
the voxel size is set to 0.02m for ScanNetV2 and S3DIS,
and to 0.3m for STPLS3D. For the augmentation during training, we use random cropping for each scene with
a maximum number of 250,000 points. During testing, a whole
scene is used as an input to the network.

Our backbone is similar to\cite{vu2022softgroup,ngo2023isbnet}, which outputs features $F_{1}$ with hidden dimension$D$ as 32 channels. For the voting, we use two set-abstraction~\cite{qi2017pointnet++,qi2019deep} layers with the ball query radius 0.2 and 0.4, respectively. The number seeds and votes are set to 1024 and 256, respectively. The number of neighbors is set to 32 for both layers, similar to~\cite{ngo2023isbnet}. For \textit{Point Migration Head}, we use two layers of dynamic convolution\cite{he2021dyco3d}, and their hidden dimensions are set to 32.  $\lambda$1, $\lambda$2, $\lambda$3, and $\lambda$4 are set to 0.5, 0.5, 1, and 1, respectively. For training and inference, we set $N_{\theta}$ and  $N_{\varphi}$ to 5 and 5, respectively. During inference, Non-Maximum-Suppression is applied to the $K$ binary masks to delete redundant masks using a confidence score 0.2 as a threshold. Following~\cite{liang2021instance, wu20223d, ngo2023isbnet, Lu_2023_ICCV}, we aggregate superpoints\cite{landrieu2018large,landrieu2019point} to align the final prediction masks on the ScanNetV2 dataset. Other details, such as the architecture of MLPs and runtime analysis, are included in the supplementary material.

\begin{figure*}[h!]
	
	\includegraphics[height=3.7cm,width=17.2cm]{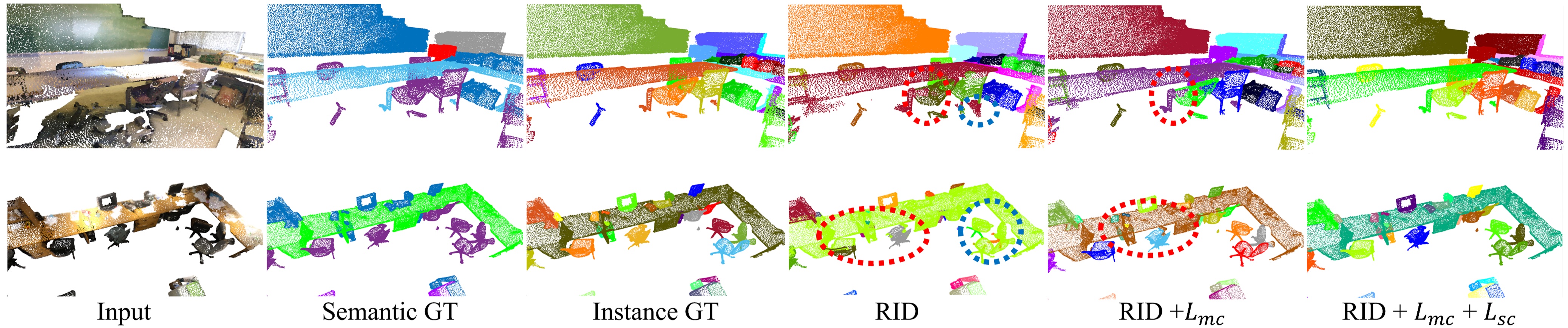}
	\caption{Visually comparing impacts of RID, $L_{mc}$, and $L_{sh}$.} 
        \vspace{-3mm}
	\label{fig:qual_ablation}
	%\vspace{-\baselineskip}
\end{figure*}

\subsection{Main Results}

\noindent \textbf{ScanNetV2}
Table~\ref{tab:scannet_test} and Table~\ref{tab:scannetv2_val} show the quantitative result of instance segmentation on the 
test and validation sets. Our proposed method achieves the highest $\text{AP}$ and $\text{AP}_{\text{50}}$ surpassing the previous strongest method by the margin of $4.1\%$ and $2.4\%$ for the test set, and  $6.7\%$ and $5.7\%$ for the validation set, respectively. Compared to the previous methods based on the coarse-to-fine approach, our method achieves $24.1\%$ of the improvement in $\text{AP}$ on the test set. In particular, for the test set, our method outperforms existing methods on instances that are typically located close to each other, such as pictures, desks, and bookshelves. This suggests that explicitly penalizing misclassified points around the edges of instances is helpful in RPM. Please refer to the supplementary material for more results about this. 

%We found a pattern of certain instances being located very close($<$ 20 cm) to each other, which can impose challenges in sharply segregating boundaries. For example, the top 3 categories found to have the same type of instances around the most are \textit{Picture}, \textit{Desk}, and \textit{Bookshelf}. On average, a picture hangs with 2.05 pictures on a wall, a desk has 1.07 adjacent desks, usually in an office scene, and a bookshelf also has 1.29 bookshelves very close, usually in a library scene. To further understand this, we plotted mAP conditioned on the number of close instances belonging to the same categories. The figure shows the drop in performance as the number of adjacent instances increases for all the methods. We think this is why we observe the performance drop in Table 1 of the main paper across all the methods. However, our method performs better as $L_{\text{mc}}$ explicitly considers misclassified points around the edges of instances in RPM without considering \hspace{-0.5mm} immense background points as other methods. 

\noindent \textbf{S3DIS}
Table~\ref{tab:s3dis} illustrates the quantitative result on Area 5. Our proposed method outperforms the second best performing method with margins of 3.3 and 2.9 in $\text{mAP}$, $\text{AP}_{\text{50}}$, and $\text{mRec}_{\text{50}}$, improving the performance of SOTA $5.7\%$, $4.2\%$, and $5.6\%$ respectively.

\noindent \textbf{STPLS3D}
Table~\ref{tab:stpls3d} shows the quantitative comparison on the validation set of STPLS3D dataset. Our method outperforms all of the existing methods, improving SOTA performance in $\text{mAP}$ and $\text{AP}_{\text{50}}$ for 3.0 and 4.3, respectively.

\subsection{Qualitative Results}
Fig.\ref{fig:qualitative} shows visual comparisons of ISBNet\cite{ngo2023isbnet}, MAFT\cite{Lai_2023_ICCV}, and our proposed Spherical Mask for challenging instances on ScanNetV2 validation set. Spherical Mask accurately segments large instances such as wall and book shelves(row 1), curtains and a window between them(row 2), a large sofa(row 3) and a circular shape sofa(row 4).

ISBNet\cite{ngo2023isbnet} struggles to segment large instances(wall and bookshelves in row 1) and a disconnected instance(curtains and a window between them in row 2). On the other hand, MAFT\cite{Lai_2023_ICCV} shows better generalization capability for learning semantics(curtains and a window between them(row 2)). However, it struggles to segment some parts of an instance that look different, as shown in sofas(row 2,3) and oversegments the instance by considering physically further away points as the same instance(wall in row 2 and sofa in row 4), probably due to the local queries that overfit to certain semantics.

%\pagebreak
\subsection{Ablation Study}
In this section, we investigate Spherical Mask with ablation studies designed for its core components.

\def\hfillx{\hspace*{-\textwidth}\hfill}
\begin{table}[h]
    \small
    \begin{minipage}[t]{.2\textwidth}
    \begin{tabular}{p{0.3cm}p{0.3cm}p{0.3cm}}
    \hline
    Method                                                      & $\text{mAP}$           & $\text{AP}_{\text{50}}$         \\ \hline
    \multicolumn{1}{c|}{HAIS~\cite{chen2021hierarchical}}       & 35.0          & 46.7          \\
    \multicolumn{1}{c|}{PointGroup~\cite{jiang2020pointgroup}}  & 23.3          & 38.5          \\
    \multicolumn{1}{c|}{ISBNet~\cite{ngo2023isbnet}}            & 49.2          & 64.0          \\ \hline
    \multicolumn{1}{c|}{Ours}                                   & \textbf{52.2} & \textbf{68.3} \\ \hline
    \end{tabular}
    \vspace{-3mm}
    \caption{Quantitative instance segmentation results on STPLS3D}
    \label{tab:stpls3d}
    \end{minipage}
    \hfillx
    \hspace{6pt}
    \vspace{-0.5mm}
    \begin{minipage}[t]{.24\textwidth}
    \begin{tabular}{p{0.2cm}p{0.2cm}p{0.2cm}p{0.3cm}p{0.3cm}p{0.3cm}}
    \hline
      RID         & $L_{mc}$        & $L_{sc}$             & $\text{mAP}$              & $\text{AP}_{\text{50}}$                & $\text{AP}_{\text{25}}$ \\ \hline
      \checkmark  &                     &                 & 51.6            & 69.4                & 86.8     \\
      \checkmark  & \checkmark          &                 & 58.5           & 77.6                & 87.5     \\
      \checkmark  &                     & \checkmark      & 56.5     & 77.1         & 87.1     \\
      \checkmark  & \checkmark          & \checkmark      & \textbf{62.3}   & \textbf{79.9}       & \textbf{88.2}     \\ \hline
    \end{tabular}
    \vspace{-3mm}
    \caption{Impact of each component on ScanNetV2
    validation set}
    \label{tab:ablation_component}
    \end{minipage}
    \vspace{-4mm}
\end{table}

\begin{table}[]
    \small
    \begin{minipage}[t!]{.2\textwidth}
    \begin{tabular}{p{0.75cm}|p{0.5cm}p{0.5cm}p{0.5cm}}
    \hline
    $N_{\theta}$/$N_{\varphi}$  & $\text{mAP}$               & $\text{AP}_{\text{50}}$ & $\text{AP}_{\text{25}}$ \\ \hline
    1/1                 &  59.6             &  77.1             &  85.3    \\
    2/2                 &  60.0             &  77.7             &  86.1    \\
    3/3                 &  60.6             &  78.4             &  87.1    \\
    4/4                 &  61.2      &  \textbf{80.0}    &  \textbf{89.3}    \\
    5/5                 &  \textbf{62.3}    &  79.9      &  88.2    \\
    6/6                 &  60.6             &  78.5             &  88.2    \\ \hline
    \end{tabular}
    \vspace{-3mm}
    \caption{Impact of $N_{\theta}$ and $N_{\varphi}$ on ScanNetV2 validation set}
    \label{tab:ablation_rays}
    
    \end{minipage}
    \hfillx
    %\hspace{15pt}
    \begin{minipage}[t]{.23\textwidth}

    \vspace{-20.9mm}
    \begin{tabular}{c|p{0.3cm}p{0.3cm}p{0.3cm}}
    \hline
    Seed/Vote & $\text{mAP}$   & $\text{AP}_{\text{50}}$   & $\text{AP}_{\text{25}}$  \\ \hline
    1024/128  & \textbf{62.3}  & 79.2 & 87.8 \\
    1024/256  & \textbf{62.3} & \textbf{79.9} & \textbf{88.2} \\
    1024/512  & \textbf{62.3}  & 79.4  & 87.6  \\
    2048/128  & 62.0  & 79.7  & 88.0  \\
    2048/256  & 61.8 & 79.6 & 87.9 \\
    2048/512  & 61.7 & 79.1 & 87.5 \\ \hline
    \end{tabular}
    \vspace{-3mm}
    \caption{Impact of seed and vote numbers on ScanNetV2
    validation set}
    \label{tab:ablation_voting}
    \end{minipage}
    \vspace{-6mm}
\end{table}

\noindent \textbf{Impact of $\theta$ and $\varphi$} is shown at Table \ref{tab:ablation_rays}. For this experiment, we fixed the backbone and the RPM with $L_{mc}$ and $L_{sc}$, and change $N_{\theta}$ and $N_{\varphi}$. In theory, increasing $N_{\theta}$ and $N_{\varphi}$ should always produce better results. However, in practice, increasing them faces an issue of high complexity, as can be seen. Using 4/4 and 5/5 of $N_{\theta}$ and $N_{\varphi}$ improves the mAP for 0.9$\%$ and 1.7$\%$, respectively. However, increasing them from 5/5 to 6/6 results in 2.8$\%$ of the performance drop, suggesting that too large $N_{\theta}$ and $N_{\varphi}$ make a negative impact on the performance due to increased complexity. For example, too large $N_{\theta}$ and $N_{\varphi}$ would result in many sectors without any foreground points, leading to biased target values for learning. Despite the variation, all configurations of $N_{\theta}$ and $N_{\varphi}$ still outperform existing methods in Table~\ref{tab:scannetv2_val}, implying that RID with RPM always improves the performance.

%\pagebreak
\noindent \textbf{Impact of Radial Point Migration} is illustrated in Table \ref{tab:ablation_component}. Here, we investigate the impact of the RPM and each loss for it. The baseline is the model trained with only RID, excluding RPM. As the baseline model cannot refine the false positive points inside radial predictions or false negative points outside, as shown in Figure~\ref{fig:qual_ablation} (parts of chairs included as tables), its performance with high iou thresholds($\text{AP}$, $\text{AP}_{\text{50}}$) are 17.1$\%$ and 13.1$\%$ worse than the full model. Including RPM with $L_{mc}$ leads 13.3$\%$ of improvement in AP, suggesting refinement to push and pull misclassified points is crucial for the performance. 
$L_{sc}$ shows 9.4$\%$ improvement in AP from the baseline, suggesting that focusing on true positive samples also contributes significantly to learning fine granularity and common features shared within an instance. As shown in Figure~\ref{fig:qual_ablation}(column 5,6), adding $L_{sc}$ improves the segmentation around the center of the objects, which was expected as the true positive samples near the instance centers could be neglected from $L_{mc}$ as they are usually inside rays. As $L_{mc}$ and $L_{sc}$ target different samples, the full model combining both $L_{mc}$ and $L_{sc}$ shows 20.7$\%$ and 15.1$\%$ improvements for $\text{AP}$ and $\text{AP}_{\text{50}}$, demonstrating the synergy of the two losses. 

\noindent \textbf{Impact of Voting Parameters} is shown in Table \ref{tab:ablation_voting}. In this experiment, we change the seed and vote numbers inside the backbone while fixing RID and RPM. As can be seen, a seed number of 1024 produces more reliable results without variation than 2048, suggesting too many seed points negatively impact the system. Despite the slight difference, we observe that changing the seed and vote produces comparably small variations of $\pm$0.6 in mAP for all settings. 

%\noindent \textbf{Impact of Encoder}

%\noindent \textbf{Runtime Analysis}

\section{Conclusion}
We present Spherical Mask, a novel coarse-to-fine approach for 3D instance segmentation in point cloud. As a coarse detection, the RID module finds instances as 3D polygons defined with center and rays. In contrast to existing coarse-to-fine approaches, the RPM module uses the polygons as soft references and migrates points efficiently in spherical coordinates to acquire final masks. We demonstrate how each module contributes to the instance segmentation and achieves state-of-the-art performances on public benchmarks ScanNet-V2, S3DIS, and STPLS3D.  

\vspace{-2mm}
\section*{Acknowledgement}
%\raggedright 
%This research was supported by ACE-OPS: From Autonomy to Cognitive assistance in Emergency OPerationS project (EP/S030832/1).
%\setstretch{0} 
\vspace{-1mm}
\raggedright This research was supported by ACE-OPS project (EP/S030832/1).
%\setstretch{0}
%This researc
%This research was supported by From Autonomy to Cognitive Assistance in Emergency OPerationS (ACE-OPS) project (EP/S030832/1).

{
    \small
    \bibliographystyle{ieeenat_fullname}
    \bibliography{main}
}


\end{document}